\documentclass[10pt,twocolumn,letterpaper]{article}

\usepackage{cvpr}              % To produce the CAMERA-READY version
\usepackage[table]{xcolor}
\usepackage{multirow}
\usepackage{utfsym}
 \usepackage{pifont}
\usepackage{wasysym}

\usepackage[utf8]{inputenc}
\usepackage[linesnumbered,ruled,vlined]{algorithm2e}
\usepackage{microtype} % For better typography
\usepackage{bm}	 % For bold math symbols
\usepackage[svgnames]{xcolor}
\definecolor{cvprblue}{rgb}{0.21,0.49,0.74}
\usepackage[pagebackref,breaklinks,colorlinks,allcolors=cvprblue]{hyperref}

\def\paperID{3096} % *** Enter the Paper ID here
\def\confName{CVPR}
\def\confYear{2026}

\title{Boosting Quantitive and Spatial Awareness for Zero-Shot Object Counting}

\author{
	Da Zhang$^{1,2*}$ \quad
	Bingyu Li$^{2,3}$ \quad
	Feiyu Wang$^{2,4}$ \quad
	Zhiyuan Zhao$^{2}$ \quad
	Junyu Gao$^{1,2\dagger}$\\
	$^{1}$Northwestern Polytechnical University \quad
	$^{2}$Institute of Artificial Intelligence (TeleAI), China Telecom \\
	$^{3}$University of Science and Technology of China \quad
	$^{4}$Fudan University \\
	{\tt\small \{dazhang900, tuzixini, gjy3035\}@gmail.com}, \\ {\tt\small libingyu0205@mail.ustc.edu.cn},  
	{\tt\small wangfy25@m.fudan.edu.cn}
}

\begin{document}
\maketitle

\begingroup
\renewcommand{\thefootnote}{\fnsymbol{footnote}}
\setcounter{footnote}{1}
\footnotetext{Work done during an internship at TeleAI.}
\setcounter{footnote}{2}
\footnotetext{Corresponding author.}
\endgroup

\begin{abstract}
	
Zero-shot object counting (ZSOC) aims to enumerate objects of arbitrary categories specified by text descriptions without requiring visual exemplars. However, existing methods often treat counting as a coarse retrieval task, suffering from a lack of fine-grained quantity awareness. 
Furthermore, they frequently exhibit spatial insensitivity and degraded generalization due to feature space distortion during model adaptation.
To address these challenges, we present \textbf{QICA}, a novel framework that synergizes \underline{q}uantity percept\underline{i}on with robust spatial \underline{c}ast \underline{a}ggregation. Specifically, we introduce a Synergistic Prompting Strategy (\textbf{SPS}) that adapts vision and language encoders through numerically conditioned prompts, bridging the gap between semantic recognition and quantitative reasoning. 
To mitigate feature distortion, we propose a Cost Aggregation Decoder (\textbf{CAD}) that operates directly on vision-text similarity maps. 
By refining these maps through spatial aggregation, CAD prevents overfitting while preserving zero-shot transferability. Additionally, a multi-level quantity alignment loss ($\mathcal{L}_{MQA}$) is employed to enforce numerical consistency across the entire pipeline. Extensive experiments on FSC-147 demonstrate competitive performance, while zero-shot evaluation on CARPK and ShanghaiTech-A validates superior generalization to unseen domains. Code will be \href{https://github.com/zhangda1018/QICA}{available}.

\end{abstract}

\section{Introduction}
\label{sec:intro}

Object counting is a fundamental computer vision task with wide ranging applications including urban surveillance \cite{ma2025scene}, ecological monitoring \cite{qazi2024animalformer, zhang2026cross}, and inventory management \cite{binyamin2025make}.
Traditional methods often focused on specific object categories requiring extensive annotated datasets for each class, limiting their scalability and adaptability to unseen scenarios \cite{wang2020nwpu, gao2025survey, lin2025point, gao2025dynamic}.
To address this inflexibility, recent research has explored class agnostic counting paradigms \cite{lu2018class, ranjan2022exemplar}. 
Among these, zero shot object counting (ZSOC) \cite{xu2023zero} presents a highly desirable goal aiming to count objects of any category specified by natural language text descriptions without needing category specific training data or visual exemplars during inference \cite{pelhan2024dave, huang2024point, wang2024language}.

\begin{figure}[t]
	\centering
	\includegraphics[width=0.48\textwidth]{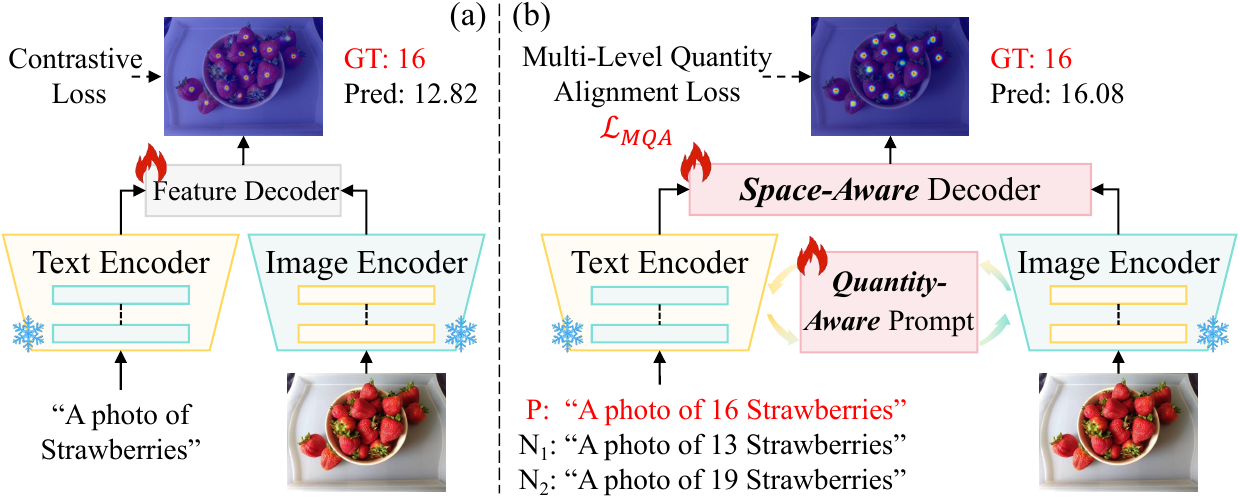} % LaTeX 自带示例图
	\caption{A \textbf{comparison} between (a) standard ZSOC and (b) QICA. (a) Existing methods typically rely on unimodal prompting and direct feature interaction, suffering from a lack of \textbf{fine-grained quantity awareness} and \textbf{spatial insensitivity}, which often leads to overfitting. (b) QICA introduces numerically conditional collaborative prompts and achieves accurate density estimation through cost aggregation decoding via $\mathcal{L}_{MQA}$.}
	\label{fig1}
\end{figure}

The advent of large-scale vision-language models (VLMs) pretrained on massive image-text pairs has significantly advanced ZSOC capabilities \cite{liang2023crowdclip, kang2024vlcounter, dai2024referring, mondal2025omnicount}. These models learn a joint embedding space that enables associating visual content with textual descriptions \cite{amini2024countgd, hui2024class, zhang2024integrating}. 
Current ZSOC methods typically employ a VLM encoder to extract image features and text embeddings, compute vision-language similarity maps to roughly locate potential object regions, and process these features through a downstream decoder for density map prediction \cite{lin2024fixed, qian2025t2icount}. 
While this paradigm has shown promising results, it faces fundamental challenges stemming from the gap between image-level VLM pretraining and pixel-level density estimation requirements \cite{wang2025exploring, zhang2025kaid}.

A critical yet often overlooked limitation is the \textbf{lack of fine-grained quantity awareness} in existing text-prompted counting methods \cite{peng2024synthesize, shi2025text}. 
While text prompts effectively specify target categories, they typically do not provide supervision regarding object quantities \cite{zhang2025enhancing}. 
Current models are primarily trained to align visual features with categorical semantics, essentially learning what an object looks like but not what constitutes different quantities visually \cite{zeng2025yolo}. 
Text prompts usually specify only object categories without incorporating numerical context, resulting in encoders that excel at recognizing what objects are but struggle to understand how many exist \cite{guo2025your, paiss2023teaching, du2024teach}. 
This quantity blindness hinders the ability to accurately distinguish between scenes containing varying numbers of objects, limiting precision especially in dense scenarios where understanding numerical differences is crucial \cite{qharabagh2024lvlm, gao2021domain}.

Another challenge arises during adaptation of pretrained VLMs to counting tasks \cite{radford2021learning, gao2025combining, rong2026unicbench}. 
Directly fine-tuning VLM encoders within conventional feature aggregation frameworks often leads to severe overfitting on training categories \cite{jia2022visual, zhou2022learning, khattak2023maple, fu2025hidden}. 
This fine-tuning process can distort the pretrained feature space, critically impairing generalization to unseen object classes during zero-shot inference. 
Existing adaptation strategies that operate on intermediate feature representations risk projecting embeddings into task-specific spaces that deviate from the original pretrained manifold \cite{wortsman2022robust}. 
Furthermore, present architectures often exhibit \textbf{spatial insensitivity} due to coarse aggregation mechanisms that neglect fine-grained details. 
Together with feature space distortion, this forces a compromise: current ZSOC approaches often resort to freezing VLM encoders entirely, sacrificing the potential benefits of adapting powerful representations to counting-specific requirements \cite{liu2025countse, li2025dcount, nguyen2025region}.

To overcome these limitations, we introduce QICA, a novel framework for ZSOC based on quantity-aware synergistic prompting and cost spatial aggregation (as shown in Figure \ref{fig1}). 
To instill quantity awareness, we propose a \textbf{synergistic prompting strategy (SPS)} that conditions learnable prompts in both text and vision encoders on numerical information. 
Unlike conventional approaches that treat branches independently, our prompts are linked via a coupling function enabling bidirectional gradient flow and joint adaptation toward quantity understanding. 
During training, we dynamically generate quantity-oriented text prompts with factual and counterfactual numerical values to teach fine-grained quantity distinction. 
To address the fine-tuning challenge, we introduce a \textbf{cost aggregation decoder (CAD)} that operates directly on vision-text similarity maps. 
By refining similarity maps through robust spatial aggregation, our approach mitigates overfitting while preserving generalization to unseen classes. 
Furthermore, we propose a \textbf{multi-level quantity alignment loss} ($\mathcal{L}_{MQA}$) that enforces numerical consistency across encoder and decoder stages through ranking constraints on embeddings and auxiliary losses.

Extensive experiments demonstrate the effectiveness of our approach. QICA achieves competitive performance on FSC-147, significantly outperforming existing text-prompted counting methods. Cross-dataset evaluation on CARPK and ShanghaiTech-A further validates strong generalization capability to unseen domains without any fine-tuning. In summary, we make the following contributions:

\begin{itemize}
	\item We propose the SPS to instill fine-grained quantity awareness. By leveraging explicit numerical conditioning and bidirectional gradient coupling, it achieves effective joint adaptation of vision and language encoders.
	\item To address spatial insensitivity and prevent feature space distortion, we introduce the CAD. It operates directly on similarity maps to preserve spatial structure while enabling robust fine-tuning.
	\item We introduce $\mathcal{L}_{MQA}$ that enforces strict numerical consistency across both encoder and decoder stages through ranking constraints and auxiliary supervision.
	\item Experiments on FSC-147 and cross-dataset validation confirm the effectiveness of QICA.
\end{itemize}

\begin{figure*}[t]
	\centering
	\includegraphics[width=1\textwidth]{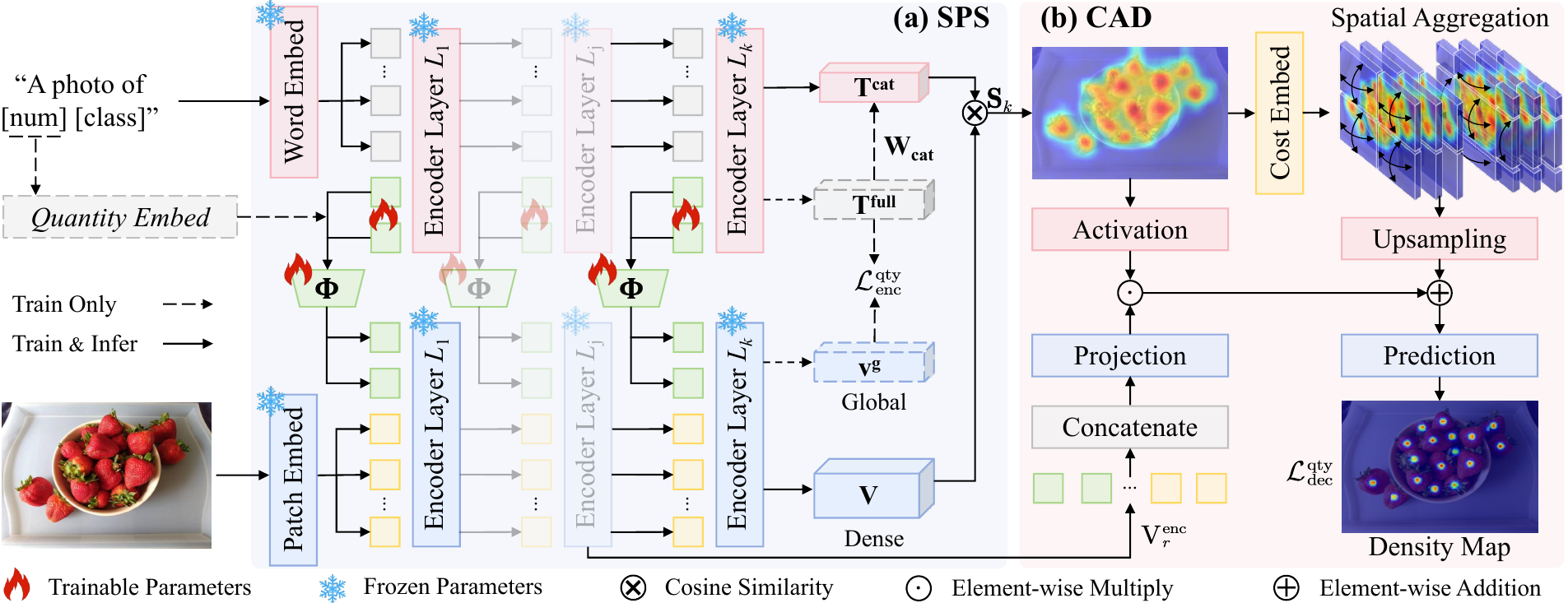} % LaTeX 自带示例图
	\caption{\textbf{Overall architecture of} \textbf{QICA}.
		(a) \textbf{SPS} jointly adapts the frozen vision and language encoders by mapping quantity-aware text prompts to visual prompts via a coupling function ($\Phi$).
		(b) \textbf{CAD} first computes a similarity map between dense visual features ($\text{V}$) and category-only text embeddings ($\text{T}^{cat}$), then refines this map through spatial aggregation and multi-scale upsampling to predict the final density map.
		The entire framework is supervised by $\mathcal{L}_{MQA}$.
		Notably, the pathways involving explicit quantity information (Quantity Embed and $\text{T}^{full}$)  are active only during training, while the model uses only category information at inference to ensure zero-shot generalization.}
		\vspace{-0.1cm}
	\label{fig2}

\end{figure*}

\section{Related Work}

\subsection{Class-Agnostic Object Counting}
Early object counting research focused on category-specific scenarios where models were trained to count particular object types such as crowds \cite{wang2020nwpu, ma2025scene, shu2022crowd, liu2023point}, vehicles \cite{hsieh2017drone, cheng2022rethinking, damiano2024can}, or cells \cite{xie2018microscopy, hatton2023human}. 
These specialized approaches achieved strong performance within their target domains but required extensive retraining for new categories \cite{cheng2022rethinking}. 
Class-agnostic counting emerged to address this limitation by formulating the task as a similarity matching problem between image regions and provided exemplars \cite{lu2018class, ranjan2022exemplar}. 
Pioneering works introduced architectures that compute feature correspondences between query images and reference patches to localize and count target objects \cite{yang2021class}. 
Subsequent methods enhanced this paradigm through improved feature extraction with advanced backbones, refined matching mechanisms via multi-scale feature fusion, and sophisticated similarity metric learning \cite{jiang2019crowd, fan2020few, xu2023zero}. 
While these exemplar-based approaches demonstrate impressive generalization across diverse object categories, they inherently require manual annotation of reference patches during inference, limiting their practical applicability in dynamic scenarios \cite{lu2018class, ranjan2021learning, shi2022represent, liu2022countr, Dukic_2023_ICCV, lin2024fixed}. Our QICA framework eliminates this dependency by leveraging text descriptions rather than visual exemplars for category specification.

\subsection{Text-Prompted Zero-Shot Counting}
The advent of large-scale VLMs has enabled zero-shot object counting through natural language specification \cite{radford2021learning, liang2023crowdclip}. 
Early approaches in this direction employed text-conditioned generative models to synthesize visual exemplars from text descriptions, which were then used in conventional exemplar-based counting frameworks \cite{wang2024enhancing, doubinsky2024semantic, zhu2024zero}. 
More recent methods directly leverage pretrained vision-language models to extract joint image-text representations \cite{jiang2023clip}.
These approaches typically compute similarity between dense image features and text embeddings to generate localization maps, which are subsequently processed by CNN-based decoders for density prediction \cite{kang2024vlcounter}. 
Various techniques have been proposed to enhance this paradigm, including hierarchical feature interaction modules, semantic-conditioned prompt tuning for visual encoders, and multi-stage architectures combining region proposal networks with vision-language matching \cite{amini2024countgd, zhang2025enhancing, qian2025t2icount}. 
However, these methods predominantly focus on categorical alignment while treating quantity information implicitly. Recent work has begun exploring numerical information in text prompts but applies uniform treatment to different quantity values without capturing fine-grained numerical relationships \cite{guo2025your, paiss2023teaching, du2024teach}. 
Furthermore, most existing approaches freeze vision-language encoders to avoid overfitting, sacrificing adaptation potential \cite{pelhan2024dave, liu2025countse}. 
In contrast, QICA introduces quantity-conditioned prompting with explicit numerical supervision through ranking constraints and employs cost aggregation on similarity maps to enable robust encoder fine-tuning without compromising generalization.

\section{Methodology}

\subsection{Problem Formulation}

Given an input image $I \in \mathbb{R}^{H \times W \times 3}$ and a text description $T$ specifying the target object category, our goal is to predict a density map $D \in \mathbb{R}^{H \times W}$ where the integral of $D$ corresponds to the total count of specified objects. During training, we have access to a dataset $\mathcal{D}_{\text{train}} = \{(I_i, T_i, D_i^{\text{GT}})\}_{i=1}^N$ containing images with corresponding text prompts and ground truth density maps. 
At inference time, the model receives novel images with text descriptions of potentially unseen categories and must predict accurate density maps without access to visual exemplars or category-specific training. The proposed QICA architecture is shown in Figure \ref{fig2}.

\subsection{Synergistic Prompting Strategy}
\label{section3.2}
SPS adapts both vision and language encoders of a pretrained VLM through learnable tokens explicitly conditioned on numerical information. 
Our approach builds upon the observation that standard prompt tuning treats vision and language branches independently, limiting their joint adaptation for quantity-aware tasks \cite{zhou2022learning, khattak2023maple, guo2025your}.

To incorporate numerical information into the prompting process, we first transform discrete quantity values into continuous representations. Given a quantity value $q$, we obtain its embedding through an embedding layer followed by linear projection, yielding $\boldsymbol{\epsilon}_q \in \mathbb{R}^{d_t}$ where $d_t$ denotes the text encoder dimension. 
During training, we generate multiple quantity values including the ground truth count $n^{\text{gt}}$ and counterfactual values $\{n^{\text{gt}} \pm k\delta\}$ where $\delta$ is dynamically determined based on $n^{\text{gt}}$ following interval-based binning strategy (see Appendix). 
This produces $K$ quantity embeddings $\{\boldsymbol{\epsilon}_{q_0}, \boldsymbol{\epsilon}_{q_1}, \ldots, \boldsymbol{\epsilon}_{q_{K-1}}\}$.

For each layer $j$ in the first $L$ layers of the text encoder, we introduce learnable prompt tokens $\boldsymbol{\Pi}^j = [\boldsymbol{\pi}_1^j, \boldsymbol{\pi}_2^j, \ldots, \boldsymbol{\pi}_m^j] \in \mathbb{R}^{m \times d_t}$ where $m$ denotes the prompt length. To condition these prompts on quantity information, we compute quantity-aware prompts as
\begin{equation}
	\hat{\boldsymbol{\Pi}}^j_k = \boldsymbol{\Pi}^j + \boldsymbol{\epsilon}_{q_k} \mathbf{1}^T
\end{equation}
where $\mathbf{1} \in \mathbb{R}^m$ is a vector of ones and $k$ indexes the quantity hypothesis. The outer product broadcasts the quantity embedding across all prompt positions. These conditioned prompts are prepended to word embeddings of text descriptions, forming an input sequence that is processed through the text encoder layers.

Following the principle of maintaining training-inference consistency, the text encoder produces two types of embeddings during training. The complete embedding $\mathbf{T}^{\text{full}}_k$ extracted from the final token output captures both categorical and numerical semantics from text descriptions containing numerical values. The category embedding $\mathbf{T}^{\text{cat}}_k$ is obtained by removing the influence of quantity tokens through a learned linear projection $\mathbf{W}_{\text{cat}}$, formally expressed as
\begin{equation}
	\mathbf{T}^{\text{cat}}_k = \mathbf{W}_{\text{cat}}(\mathbf{T}^{\text{full}}_k)
\end{equation}
At inference time, input text contains only category information without numerical values. To ensure training-inference consistency, we employ a dual-path strategy: during training, quantity-aware prompts generate complete embeddings $\mathbf{T}^{\text{full}}$, which are then projected through $\mathbf{W}_{\text{cat}}$ to extract category-only semantics $\mathbf{T}^{\text{cat}}$. During inference, category-only prompts naturally produce embeddings with identical semantic content to $\mathbf{T}^{\text{cat}}$, bypassing the need for $\mathbf{W}_{\text{cat}}$ projection. 
This design ensures that similarity computation receives semantically equivalent category embeddings in both phases, maintaining consistent decoder inputs despite different computational pathways.

To establish synergistic adaptation between modalities, we introduce coupling functions $\{\boldsymbol{\Phi}^j\}_{j=1}^L$ that project language prompts to vision prompts as
\begin{equation}
	\boldsymbol{\Psi}^j_k = \boldsymbol{\Phi}^j(\hat{\boldsymbol{\Pi}}^j_k)
\end{equation}
where $\boldsymbol{\Phi}^j$ is implemented as a linear projection from $d_t$ to $d_v$. These vision prompts are inserted into corresponding layers of the vision encoder alongside image patch embeddings. This coupling mechanism enables bidirectional gradient flow where updates to language prompts influence vision prompts and vice versa, facilitating joint adaptation of both encoders toward quantity-sensitive feature extraction. The vision encoder processes the image to produce dense visual embeddings $\mathbf{V} \in \mathbb{R}^{(h \times w) \times d_v}$ and global image representation $\mathbf{v}^g \in \mathbb{R}^{d_v}$ from the class token. 
Crucially, through the quantity alignment objective described later (Section \ref{section3.4}), the encoder learns to implicitly encode quantity information within visual features, enabling accurate counting even when numerical supervision is unavailable at inference.

\subsection{Cost Aggregation Decoder}

Compared to previous work \cite{jiang2023clip, kang2024vlcounter, qian2025t2icount}, CAD operates directly on vision-text similarity maps rather than intermediate feature. 
This cost-based paradigm maintains the integrity of the pretrained embedding space while enabling robust fine-tuning \cite{hu2022lora}. 
Critically, to ensure training-inference consistency (Section \ref{section3.2}), we compute similarity maps using category embeddings $\mathbf{T}^{\text{cat}}_k$ rather than $\mathbf{T}^{\text{full}}_k$.

For each quantity hypothesis $k$, we compute the cosine similarity between dense visual embeddings and the corresponding category embedding as
\begin{equation}
	\mathbf{S}_k(i) = \frac{\mathbf{V}(i) \cdot \mathbf{T}^{\text{cat}}_k}{\|\mathbf{V}(i)\| \|\mathbf{T}^{\text{cat}}_k\|}
\end{equation}
where $i$ denotes spatial position and $\mathbf{S}_k \in \mathbb{R}^{h \times w}$ represents the similarity map. This design ensures that both training and inference compute similarities based solely on categorical matching, with quantity information implicitly encoded in visual features. We embed this similarity map into a higher-dimensional feature space through a convolutional layer, obtaining $\mathbf{G}_k \in \mathbb{R}^{h \times w \times d_g}$ where $d_g$ is the embedding dimension.

\begin{figure}[t]
	\centering
	\includegraphics[width=0.48\textwidth]{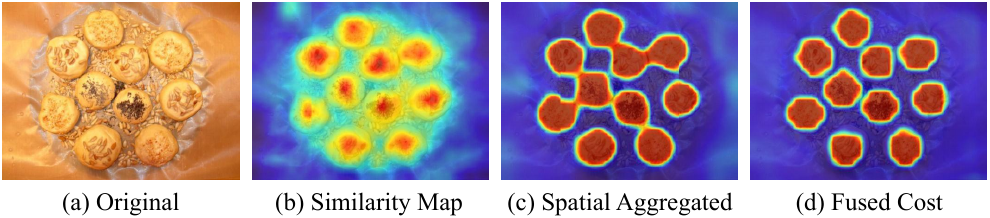} % LaTeX 自带示例图
	\caption{\textbf{Visualization of the CAD pipeline}. 
		(a) Original image. 
		(b) The similarity map derived from the fine-tuned CLIP exhibits \textbf{fine-grained quantity awareness}, effectively highlighting target instances while suppressing noise. 
		(c) The aggregated cost map and (d) the final fused map demonstrate the recovery of \textbf{fine-grained spatial structure}, transforming coarse activations into precise density predictions.
		}
	\label{fig3}

\end{figure}

The embedded similarity features are refined through spatial aggregation that leverages image structure to enhance localization. We employ Swin Transformer blocks for their computational efficiency and global receptive fields, operating independently on each quantity hypothesis as
\begin{equation}
	\mathbf{G}'_k = \mathcal{A}^{\text{spa}}(\mathbf{G}_k; \mathbf{V})
\end{equation}
where $\mathcal{A}^{\text{spa}}$ denotes the spatial aggregation module consisting of consecutive Swin Transformer blocks with windowed and shifted-window attention patterns.
Inspired by \cite{sun2018pwc, cho2024cat}, we optionally provide visual embeddings $\mathbf{V}$ as guidance to the attention computation through linear projection and concatenation to query and key features. 
This aggregation refines coarse similarity maps by exploiting spatial coherence, suppressing isolated activations while enhancing connected regions corresponding to actual objects.

To recover high-resolution density predictions, we progressively upsample the aggregated similarity features while incorporating multi-scale information from the vision encoder through skip connections. At each upsampling stage, we extract intermediate features from the vision encoder and modulate these connections by similarity maps to emphasize quantity-relevant regions, formulated as
\begin{equation}
	\mathbf{G}^{(r+1)} = \text{Conv}\big(\text{Up}(\mathbf{G}^{(r)}) + \mathbf{W}_{\text{proj}}(\mathbf{V}^{\text{enc}}_r) \odot \sigma(\mathbf{S}_k)\big)
\end{equation}
where $\mathbf{V}^{\text{enc}}_r$ denotes encoder features from stage $r$, $\mathbf{W}_{\text{proj}}$ is a projection layer, and $\sigma$ represents sigmoid activation. The final density map $\hat{D}_k$ for quantity hypothesis $k$ is obtained through a prediction head applied to the upsampled features. Figure \ref{fig3} visually illustrates the entire polymerization process. For detailed illustrations of the architecture for each component, please refer to the supplementary materials.

\subsection{Multi-level Quantity Alignment Loss}
\label{section3.4}
We introduce $\mathcal{L}_{\text{MQA}}$ that enforces quantity awareness across multiple stages of the network through carefully designed ranking constraints. This multi-level design creates supervision signals at both encoder and decoder stages, enabling comprehensive quantity understanding throughout the pipeline.

To teach the encoder to distinguish fine-grained quantity differences, we leverage complete embeddings $\mathbf{T}^{\text{full}}_k$ that contain numerical information. We compute cosine similarities between the global image representation and all complete text embeddings as
\begin{equation}
	\alpha_k = \frac{\mathbf{v}^g \cdot \mathbf{T}^{\text{full}}_k}{\|\mathbf{v}^g\| \|\mathbf{T}^{\text{full}}_k\|}
\end{equation}
where $\alpha_k$ is the alignment score for the $k$-th quantity hypothesis. Let $\alpha_0$ denote the similarity for the factual quantity prompt and $\{\alpha_i\}_{i=1}^{K-1}$ denote similarities for counterfactual prompts sorted by their distance from the ground truth. The encoder-level loss enforces ranking constraints as:

\begin{equation}
	\begin{gathered} % <-- 把 aligned 换成 gathered
		\mathcal{L}^{\text{qty}}_{\text{enc}} = \frac{1}{K-1}\sum_{i=1}^{K-1} f(\alpha_i - \alpha_0) + \\ % <-- 移除了行首的 &
		\frac{1}{K-3}\sum_{i=1}^{\frac{K-1}{2}-1}  [f(\alpha_i - \alpha_{i+1})  + f(\alpha_{\frac{K+1}{2}+i} - \alpha_{\frac{K+1}{2}+i+1})] % <-- 移除了行首的 &
	\end{gathered}
\end{equation}
where $f(\cdot)$ represents the ReLU function. The first term ensures the factual prompt achieves highest similarity, while the second enforces that prompts with counts closer to ground truth yield higher similarities. 
This ranking constraint teaches the encoder to implicitly encode quantity information in visual features by establishing correspondence between visual quantity prompts and numerical values in text.

At the decoder level, we enforce that predicted counts align with their corresponding quantity hypotheses, creating auxiliary supervision signals that fully utilize multiple density predictions. 
For each density map $\hat{D}_k$, we compute the predicted count $\hat{n}_k = \sum \hat{D}_k$ and enforce
\begin{equation}
	\mathcal{L}^{\text{qty}}_{\text{dec}} = \|\hat{n}_0 - n^{\text{gt}}\|_2^2 + \beta\sum_{k=1}^{K-1}\|\hat{n}_k - q_k\|_2^2
\end{equation}
where $q_k$ represents the quantity value in the $k$-th prompt and $\beta=0.1$ (experientially) controls the weight of auxiliary terms. 
The first term ensures accurate counting for the factual hypothesis, while the auxiliary terms encourage the decoder to learn quantity-dependent prediction patterns.

The complete training objective combines density estimation loss with multi-level quantity alignment as:
\begin{equation}
	\mathcal{L}_{\text{MQA}} = \|\hat{D}_0 - D^{\text{GT}}\|_2^2 + \lambda_1 \mathcal{L}^{\text{qty}}_{\text{enc}} + \lambda_2 \mathcal{L}^{\text{qty}}_{\text{dec}}
\end{equation}
where $\hat{D}_0$ corresponds to the factual quantity prediction and $\lambda_1=0.1$, $\lambda_2=0.05$ balance the loss terms. During training, we process each image with multiple quantity-oriented text prompts, performing independent forward passes for each hypothesis while sharing encoder parameters. This creates diverse supervision signals from a single sample, enabling quantity-sensitive learning. At inference, only a category-specific prompt is provided, and the model produces density predictions by leveraging the quantity understanding implicitly acquired during training.

\section{Experiments}

\subsection{Experimental Setup}
\textbf{Datasets:}
We conduct comprehensive experiments to validate the effectiveness of QICA on standard benchmarks for ZSOC. Primary evaluation dataset is FSC-147 \cite{ranjan2021learning}, which comprises 6,135 images spanning 147 diverse object categories. The dataset is partitioned into 3,659 training images, 1,286 validation images, and 1,190 test images.
To assess cross-domain generalization capability, we additionally evaluate on CARPK and ShanghaiTech-A \cite{zhang2016single} without any fine-tuning. 
The CARPK dataset contains 1,448 parking lot images captured from drone perspectives at approximately 40 meters altitude, collectively annotating nearly 90,000 vehicles. 
The ShanghaiTech-A dataset contains 300 train and 182 test images of dense crowd.

\begin{table*}[t]
	\centering
	\caption{\textbf{Performance comparison} of QICA with other SOTA models on FSC-147 dataset. The best performance for each scheme is highlighted in \textbf{bold}, and the second-best performance for the zero-shot setting is \underline{underlined}.} % 请替换为您的表格标题
	\label{table1}  % 用于交叉引用的标签
	% 使用 \resizebox 将表格宽度调整为 \linewidth
	\resizebox{1\linewidth}{!}{%
		\begin{tabular}{c|cccccccccc}
			\toprule
			\multirow{2}{*}{Scheme} & \multirow{2}{*}{Method} & \multirow{2}{*}{Venue} & \multirow{2}{*}{Exemplar} & \multirow{2}{*}{Backbone} & \multicolumn{2}{c}{Val} & \multicolumn{2}{c}{Test} & \multicolumn{2}{c}{Avg.} \\

			\cmidrule(lr){6-7} \cmidrule(lr){8-9} \cmidrule(lr){10-11}

			& & & & & MAE$\downarrow$ & RMSE$\downarrow$ & MAE$\downarrow$ & RMSE$\downarrow$ & MAE$\downarrow$ & RMSE$\downarrow$ \\
			\midrule

			\multirow{7}{*}{Few-shot} & FamNet \cite{ranjan2021learning} & CVPR'21 & Visual & ResNet-50 & 24.32 & 70.94 & 22.56 & 101.54 & 23.44 & 86.24 \\
			& CountTR \cite{liu2022countr} & WACV'22 & Visual & ViT-B/16 & 13.13 & 49.83 & 11.95 & 91.23 & 12.54 & 70.53 \\
			& LOCA \cite{Dukic_2023_ICCV} & ICCV'23 & Visual & ResNet-50 & 10.24 & \underline{32.56} & 10.97 & \underline{56.97} & 10.61 & \underline{44.77} \\
			& PseCo \cite{huang2024point} & CVPR'24 & Visual & SAM-ViT-H/16 & 15.31 & 68.34 & 13.05 & 112.89 & 14.18 & 90.62 \\
			& CountGD \cite{amini2024countgd} & NeurIPS'24 & Visual \& Text & GDINO-Swin-B & \textbf{7.10} & \textbf{26.08} & \textbf{6.75} & \textbf{43.65} & \textbf{6.93} & \textbf{34.87} \\
			& CountSE \cite{liu2025countse} & ICCV'25 & Text & GDINO-Swin-B & \underline{8.51} & 54.93 & 7.84 & 82.99 & \underline{8.18} & 68.96 \\
			& TMR \cite{jo2025few} & ICCV'25 & Visual & SAM-ViT-H/16 & - & - & 11.63 & 57.46 & 11.63 & 57.46 \\
			\midrule

			\multirow{12}{*}{Zero-shot} & ZSC \cite{xu2023zero} & CVPR'23 & Text & ResNet-50 & 26.93 & 88.63 & 22.09 & 115.17 & 24.51 & 101.90 \\
			& CLIP-Count \cite{jiang2023clip} & ACM MM'23 & Text & ViT-B/16 & 18.79 & 61.18 & 17.78 & 106.62 & 18.29 & 83.90 \\
			& CounTX \cite{amini2023open} & BMVC'23 & Text & ViT-B/16 & 17.76 & 65.21 & 16.70 & 105.21 & 17.23 & 85.21 \\
			& VA-Count \cite{zhu2024zero} & ECCV'24 & Text & ViT-B/16 & 17.87 & 73.22 & 17.88 & 129.31 & 17.88 & 101.27 \\
			& VLCounter \cite{kang2024vlcounter} & AAAI'24 & Text & ViT-B/16 & 18.06 & 65.13 & 17.05 & 106.16 & 17.56 & 85.65 \\
			& DAVE \cite{pelhan2024dave} & CVPR'24 & Visual & ResNet-50 & 15.48 & 52.57 & 14.90 & 103.42 & 15.19 & 78.00 \\
			& GeCo \cite{pelhan2024novel} & NeurIPS'24 & Visual & SAM-ViT-H/16 & 14.81 & 64.95 & 13.30 & 108.72 & 14.06 & 86.84 \\
			& CountGD \cite{amini2024countgd} & NeurIPS'24 & Text & GDINO-Swin-B & \textbf{12.14} & \textbf{47.51} & 14.76 & 120.42 & 13.45 & 83.97 \\
			& T2ICount \cite{qian2025t2icount} & CVPR'25 & Text & Stable Diff-v1.5 & 13.78 & 58.78 & \textbf{11.76} & 97.86 & \underline{12.77} & 78.32 \\
			& YOLO-Count \cite{zeng2025yolo} & ICCV'25 & Text & YOLOv8-L & 14.80 & 96.14 & 15.43 & \textbf{58.36} & 15.12 & \underline{77.25} \\
			\cmidrule(lr){2-11}
%			\rowcolor{gray!15} % <-- 在这里添加了 \rowcolor
			& \cellcolor{gray!15} QICA & \cellcolor{gray!15} Ours & \cellcolor{gray!15} Text & \cellcolor{gray!15} ViT-B/16 & \cellcolor{gray!15} 13.82 & \cellcolor{gray!15} 60.24 & \cellcolor{gray!15} 13.05 & \cellcolor{gray!15} 104.17 & \cellcolor{gray!15} 13.44 & \cellcolor{gray!15} 82.21 \\
%			\rowcolor{gray!15} % <-- 在这里添加了 \rowcolor
			& \cellcolor{gray!15} $\text{QICA}^{\dag}$ & \cellcolor{gray!15} Ours & \cellcolor{gray!15} Text & \cellcolor{gray!15} ViT-L/14 & \cellcolor{gray!15} \underline{12.98} & \cellcolor{gray!15} \cellcolor{gray!15} \underline{56.35} & \cellcolor{gray!15} \underline{12.41} & \cellcolor{gray!15}  \underline{97.28} & \cellcolor{gray!15} \textbf{12.70} & \cellcolor{gray!15} \textbf{76.82} \\
			\bottomrule
		\end{tabular}%
	} 

\end{table*}

\noindent
\textbf{Implementation Details:}
We implement QICA with two backbone configurations: CLIP ViT-B/16 and ViT-L/14, leveraging stronger pretrained representations with minimal additional trainable parameters. 
Both CLIP encoders remain frozen during training, with adaptation via learnable prompts in the first 9 layers (length$=2$). 
Text encoder uses 12 layers with dimension 512, while vision encoders vary in depth (12 vs 24 layers) and dimension (768 vs 1024). 
ViT-B/16 processes $384 \times 384$ images into $24 \times 24$ features, whereas ViT-L/14 handles $392 \times 392$ images producing $28 \times 28$ features. 
The decoder employs 2 Swin Transformer blocks (dimension 128) with two-stage upsampling and adaptive skip connections (from layers 4 and 8 for ViT-B/16, layers 8 and 16 for ViT-L/14).
Training uses AdamW optimizer for 200 epochs with learning rate $1 \times 10^{-4}$ and weight decay $1 \times 10^{-2}$. 
We generate $K=5$ quantity hypotheses per image via adaptive interval binning, with loss weights $\lambda_1=0.1$ and $\lambda_2=0.05$. 
Standard augmentations are applied. Evaluation uses MAE and RMSE metrics on NVIDIA 80G A800 GPUs. Detailed information are in the Appendix.

\subsection{Comparison with Advanced Methods}

\textbf{Results on FSC-147:}
We evaluate QICA against recent methods on FSC-147 in Table \ref{table1}. 
ViT-L/14 configuration achieves 12.41 MAE and 97.28 RMSE on the test set, establishing competitive performance within the zero-shot counting paradigm. 
Compared to CLIP-based methods VLCounter, CLIP-Count, and CounTX, QICA demonstrates improvements of 27.2\%, 30.2\%, and 25.7\% respectively. 
ViT-B/16 variant attains 13.05 MAE, surpassing CountGD despite using lighter CLIP backbone, suggesting our cost aggregation effectively preserves pretrained knowledge. 
Compared to diffusion-based T2ICount, QICA remains competitive with 0.65 MAE gap while exhibiting superior generalization as evidenced by minimal validation-test variance versus CountGD text mode degradation of 2.62 MAE. 
Both configurations substantially outperform earlier zero-shot methods, validating the effectiveness of different components.

\begin{table}[t]
	\centering
	\caption{Comparison with SOTA methods on CARPK.} % 请替换为您的表格标题
	\label{table2}  % 用于交叉引用的标签
	% l = 左对齐, r = 右对齐 (r更适合数字，使小数点对齐)
	\resizebox{1\linewidth}{!}{
	\begin{tabular}{ccccc}
		\toprule
		Method & Exemplar & Fine-tuned & MAE$\downarrow$ & RMSE$\downarrow$ \\
		\midrule
		CountTR \cite{liu2022countr} & Viusal & \usym{2713} & 5.75 & 7.45 \\
		LOCA \cite{Dukic_2023_ICCV} & Viusal &\usym{2717} & 9.97 & 12.51 \\
		CountGD \cite{amini2024countgd} & Visual \& Text &\usym{2717} & 3.68 & 5.17 \\
		\midrule % 在图片中，这里有一条分隔线，用于区分不同的Exemplar类型
		CLIP-Count \cite{jiang2023clip} & Text &\usym{2717} & 11.96 & 16.61 \\
		CounTX \cite{amini2023open} & Text &\usym{2713} & 8.13 & 10.87 \\
		VA-Count \cite{zhu2024zero} & Text &\usym{2717} & 10.63 & 13.20 \\
		VLCounter \cite{kang2024vlcounter} & Text &\usym{2717} & 6.46 & 8.68 \\
		CountGD \cite{amini2024countgd} & Text &\usym{2717} & \textbf{5.98} & \textbf{7.41} \\
		T2ICount \cite{qian2025t2icount} & Text &\usym{2717} & 8.61 & 13.47 \\
		\midrule
		\cellcolor{gray!15}QICA (Ours) & \cellcolor{gray!15} Text &\cellcolor{gray!15} \usym{2717} &\cellcolor{gray!15} 6.38 & \cellcolor{gray!15}8.20 \\
		\cellcolor{gray!15} $\text{QICA}^{\dag}$ (Ours) &\cellcolor{gray!15} Text &\cellcolor{gray!15}\usym{2717} & \cellcolor{gray!15}\underline{6.07} &\cellcolor{gray!15} \underline{7.82} \\
		\bottomrule
	\end{tabular}

}

\end{table}

\noindent
\textbf{Results on CARPK:}
We evaluate cross-domain generalization on CARPK without fine-tuning as shown in Table \ref{table2}. 
QICA achieves 6.07 MAE with ViT-L/14 and 6.38 MAE with ViT-B/16, exhibiting 51\% error reduction from FSC-147. 
While CountGD attains 3.83 MAE through specialized grounding capabilities, QICA demonstrates competitive text-based performance, surpassing T2ICount and approaching DINOv2-based methods. The consistent ViT-L/14 advantage validates our architecture scalability.

\noindent
\textbf{Results on ShanghaiTech-A:}
On ShanghaiTech-A featuring extreme crowd density, QICA establishes new SOTA among open-set methods as shown in Table \ref{table3}. 
ViT-L/14 achieves 140.7 MAE and 258.3 RMSE, surpassing CountGD (141.9 MAE) despite using lighter CLIP features. ViT-B/16 attains 146.2 MAE, outperforming CLIP-Count and CounTX by 24.1\% and 33.5\% respectively. 
This validates that our cost aggregation and quantity-aware prompting effectively address dense localization challenges, with spatial aggregation proving particularly valuable in extreme counting scenarios.

\subsection{Ablation Study}
\begin{table}[t]
	\centering
	\caption{Comparison with SOTA methods including specific category counting on ShanghaiTech-A test set. * means the results are reproduced by open-source codes.} % 请替换为您的表格标题
	\label{table3}  % 用于交叉引用的标签
	% l = 左对齐 (用于文本), r = 右对齐 (用于数字，使小数点对齐)
	\resizebox{1 \linewidth}{!}{
		\begin{tabular}{		
				>{\centering\arraybackslash}p{2.5cm} % 第1列 (Method)
				>{\centering\arraybackslash}p{2cm} % 第2列 (Exemplar)
				>{\centering\arraybackslash}p{2cm}   % 第3列 (Fine-tuned)
				>{\centering\arraybackslash}p{1.5cm} % 第4列 (MAE)}
		}
		
		\toprule
		Method & Category & MAE$\downarrow$ & RMSE$\downarrow$ \\
		\midrule
		MCNN \cite{zhang2016single} & Specific & 221.4 & 357.8 \\
		CrowdClip \cite{liang2023crowdclip} & Specific & 217.0 & 322.7 \\
		\midrule
		RCC \cite{hobley2022learning}& Open-set & 240.1 & 366.9 \\
		CounTX \cite{amini2023open} & Open-set & 219.8 & 351.0 \\
		CLIP-Count \cite{jiang2023clip} & Open-set & 192.6 & 308.4 \\
		CountGD* \cite{amini2024countgd} & Open-set & \underline{141.9} & \textbf{258.0} \\
		VLPG \cite{zhai2024zero}& Open-set & 178.9 & 284.6 \\
		\midrule % 对应图片中 "VLPG" 和 "QICA" 之间的分隔线
		% 下面使用了 \textcolor{gray}{\small ...} 来满足您的特殊要求
		\cellcolor{gray!15} QICA & \cellcolor{gray!15}Open-set &\cellcolor{gray!15} 146.2 & \cellcolor{gray!15}269.6 \\
		\cellcolor{gray!15} $\text{QICA}^{\dag}$  & \cellcolor{gray!15}Open-set & \cellcolor{gray!15}\textbf{140.7} &\cellcolor{gray!15} \underline{258.3} \\
		\bottomrule
	\end{tabular}
}
\end{table}

\begin{table}[t]
	\centering
	\caption{Ablation Study on FSC-147. `TP' refers `Text Prompts only' (no quantity), `SP' represents `Synergistic Prompting'.
	Baseline/intermediate configurations (rows 1-3) process $\mathbf{V}$ directly, while CAD configurations (rows 4-5) operate on $\mathbf{V}$-$\mathbf{T}^{cat}$ similarity maps.
	} % 请替换为您的表格标题
	\label{table4}  % 用于交叉引用的标签
	% 定义列格式：
	% 4 列居中 (c) 用于符号
	% 4 列右对齐 (r) 用于数字，以便小数点对齐
	\resizebox{1 \linewidth}{!}{
		% 已将列定义从 {cccc rr rr rr} 更改为 {cccc rr rr}
		\begin{tabular}{		
				>{\centering\arraybackslash}p{0.6cm} % 第1列 (Method)
				>{\centering\arraybackslash}p{0.6cm} % 第2列 (Exemplar)
				>{\centering\arraybackslash}p{0.6cm}   % 第3列 (Fine-tuned)
				>{\centering\arraybackslash}p{0.8cm} % 第4列 (MAE)}
				>{\centering\arraybackslash}p{0.7cm}   % 第3列 (Fine-tuned)
				>{\centering\arraybackslash}p{1cm} % 第4列 (MAE)}
				>{\centering\arraybackslash}p{0.7cm}   % 第3列 (Fine-tuned)
				>{\centering\arraybackslash}p{1cm} % 第4列 (MAE)}
		}
		
			\toprule

			\multirow{2}{*}{TP} & \multirow{2}{*}{SP} & \multirow{2}{*}{CAD} & \multirow{2}{*}{$\mathcal{L}_{MQA}$} & \multicolumn{2}{c}{Val} & \multicolumn{2}{c}{Test} \\

			\cmidrule(lr){5-6} \cmidrule(lr){7-8}

			& & & & MAE$\downarrow$ & RMSE$\downarrow$ & MAE$\downarrow$ & RMSE$\downarrow$ \\
			\midrule

			\Circle & \Circle & \Circle & \Circle & 16.82 & 68.45 & 16.15 & 124.32 \\
			\CIRCLE & \Circle & \Circle & \Circle & 15.46 & 63.28 & 14.89 & 115.63 \\
			\Circle & \CIRCLE & \Circle & \Circle & 14.27 & 59.14 & 13.72 & 107.85 \\
			\Circle & \CIRCLE & \CIRCLE & \Circle & 13.41 & 57.62 & 12.88 & 99.74 \\
			\Circle & \CIRCLE & \CIRCLE & \CIRCLE & \textbf{12.98} & \textbf{56.35} & \textbf{12.41} & \textbf{97.28} \\
			\bottomrule
		\end{tabular}
	}
\end{table}

We systematically evaluate each component contribution on FSC-147 with ViT-L/14 as shown in Table \ref{table4}. 
From baseline CLIP fine-tuning (16.15 test MAE), adding text prompts (no quantity) reduces error to 14.89 MAE through learnable textual representations. 
Synergistic prompting yields 13.72 MAE with 8.0\% improvement validating cross-modal adaptation effectiveness. 
Cost aggregation decoder achieves 12.88 MAE (6.1\% gain) by preserving pretrained knowledge through similarity map operations. 
Multi-level quantity alignment reaches 12.41 MAE via joint encoder-decoder supervision. The cumulative 23.2\% improvement confirms all components work synergistically.

\begin{figure}[ht] % [ht] 是浮动选项，尝试将图片放在“这里(here)”或“顶部(top)”
	\centering % 整个 figure 环境居中
	
	% --- 第一个子图 ---
	\begin{subfigure}[b]{0.236\textwidth}
		\centering
		\includegraphics[width=\textwidth]{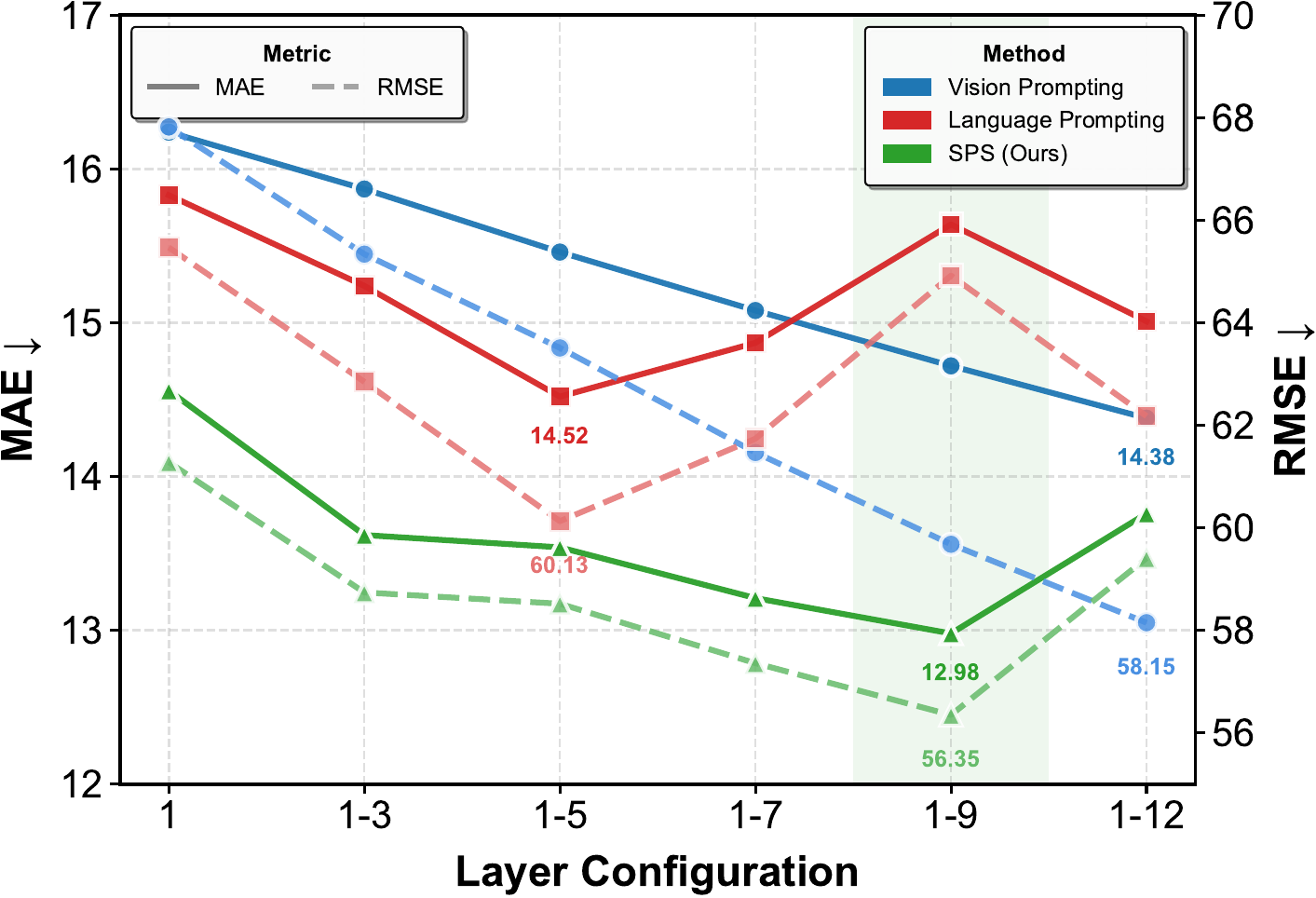} % <-- 替换为您的图片路径
		\caption{}
		\label{fig:sub1}
	\end{subfigure}
	\hfill % 这是一个“弹簧”，在两个子图之间添加水平间距
	% --- 第二个子图 ---
	\begin{subfigure}[b]{0.236\textwidth}
		\centering
		\includegraphics[width=\textwidth]{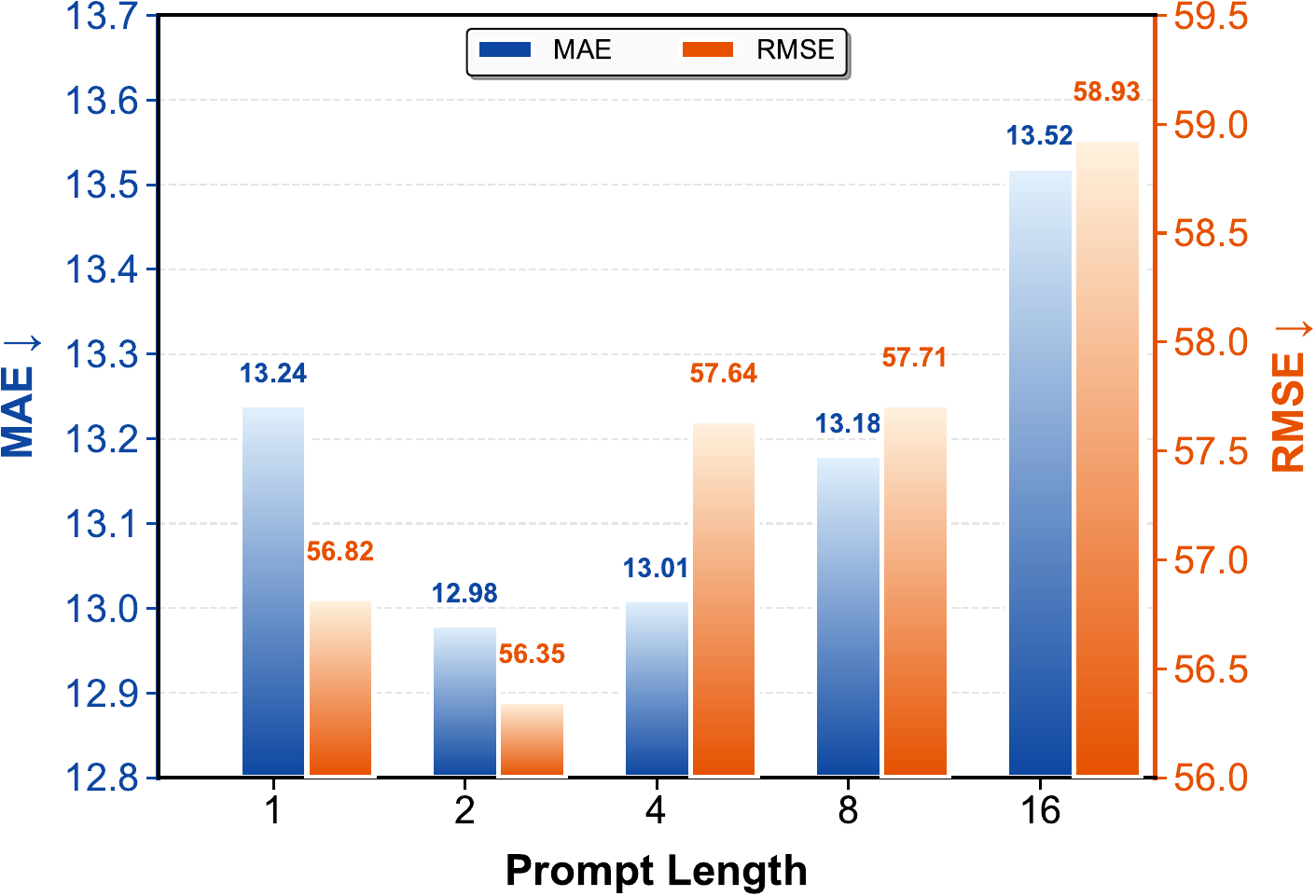} % <-- 替换为您的图片路径
		\caption{}
		\label{fig:sub2}
	\end{subfigure}
	
 % 可选：在子图和主标题之间添加一点垂直空间
	
	% --- 主标题 ---
	\caption{Ablation on prompt depth (a) and prompt length (b) in QICA. We report results on the validation sets of FSC-147.}
	\label{fig:main_figure}
\end{figure}

\section{Model Analysis}
\subsection{Analysis of Prompt}

\textbf{Prompt Depth:} 
Figure \ref{fig:sub1} shows prompt depth effects on FSC-147 validation set. Vision-only prompting improves monotonically to 14.38 MAE at 1-12 layers, while language-only peaks at 1-5 layers (14.52 MAE) before degrading. SPS achieves 12.98 MAE at 1-9 layers, outperforming both unimodal strategies by enabling complementary adaptation where quantity-aware text guides vision refinement.

\noindent
\textbf{Prompt Length:} Figure \ref{fig:sub2} illustrates length sensitivity. Optimal performance occurs at length 2 (12.98 MAE), with degradation beyond length 4. 
This indicates compact prompts sufficiently encode quantity semantics while avoiding over-parameterization, contrasting with assumptions that longer prompts provide greater expressiveness.

\begin{table}[ht]
	\centering
	 \caption{Results for different value of K on FSC-147. More results are in the appendix.}
	 \label{table5}
	\resizebox{1 \linewidth}{!}{
	\begin{tabular}{		
			>{\centering\arraybackslash}p{1cm} % 第1列 (Method)
			>{\centering\arraybackslash}p{2cm} % 第2列 (Exemplar)
			>{\centering\arraybackslash}p{1.2cm}   % 第3列 (Fine-tuned)
			>{\centering\arraybackslash}p{1.2cm} % 第4列 (MAE)}
		>{\centering\arraybackslash}p{1.2cm}   % 第3列 (Fine-tuned)
	>{\centering\arraybackslash}p{1.2cm}   % 第3列 (Fine-tuned)
}
		\toprule
		& \textbf{K} & \textbf{1} & \textbf{3} & \textbf{5} & \textbf{7} \\ 
		\midrule
		\multirow{2}{*}{Val}  & MAE$\downarrow$        & 13.76      & 13.21      & \textbf{12.98 }     & 13.12      \\  
		& RMSE$\downarrow$       & 58.24      & 57.16      & \textbf{56.35}      & 56.89      \\
		\midrule
		\multirow{2}{*}{Test} & MAE$\downarrow$        & 13.18      & 12.67      & \textbf{12.41}      & 12.58      \\  
		& RMSE$\downarrow$       & 101.45     & 98.92      & \textbf{97.28}      & 98.16      \\ 
		\bottomrule
	\end{tabular}
}
\end{table}

\noindent
\textbf{Number of Quantity K:}
Table \ref{table5} examines \textbf{K} effects on FSC-147. 
Using only factual hypothesis (K=1) yields 13.76 validation MAE, while K=5 achieves optimal performance at 12.98 validation MAE and 12.41 test MAE through symmetric counterfactual pairs. 
Further increasing to K=7 degrades performance to 13.12 MAE, indicating excessive hypotheses introduce optimization challenges without benefits.

\subsection{Effect of Loss Weight}

\begin{figure}[ht]
	\centering
	\includegraphics[width=0.48\textwidth]{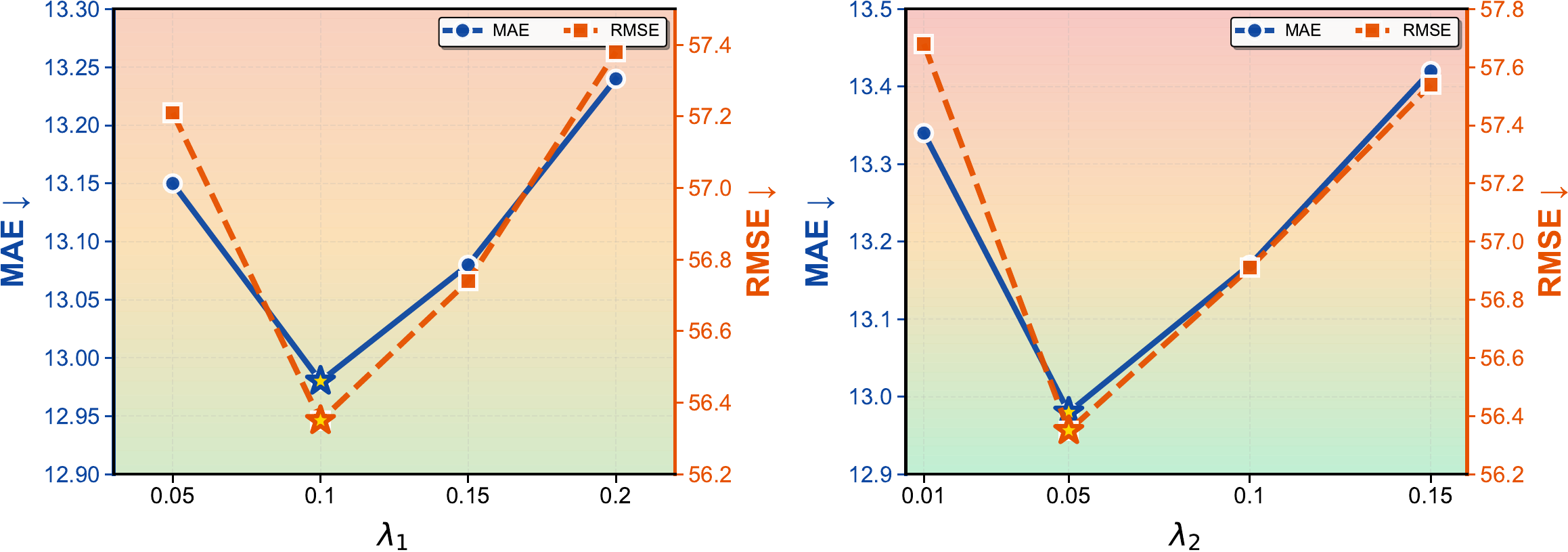} % LaTeX 自带示例图
	\caption{
		Sensitivity analysis of loss weights on FSC-147 validation set. Left shows the effect of $\lambda_1$ on $\mathcal{L}_{\text{enc}}^{\text{qty}}$ while right illustrates $\lambda_2$ on $\mathcal{L}_{\text{dec}}^{\text{qty}}$. \textcolor{Gold}{\ding{72}} Gold stars mark optimal configurations.
	}
	\label{fig5}
	\vspace{-0.1cm}
\end{figure}

Figure \ref{fig5} presents sensitivity analysis of loss weights $\lambda_1$ and $\lambda_2$ on FSC-147 validation set. For decoder quantity consistency, $\lambda_1=0.1$ achieves optimal 12.98 MAE, with deviations degrading performance up to 2.0\%. Encoder quantity alignment exhibits higher sensitivity, where $\lambda_2=0.05$ reaches optimum while excessive values (0.15) degrade by 3.4\%. This reflects that ranking constraints on frozen encoder outputs require careful calibration. 
Notably, insufficient encoder supervision ($\lambda_2=0.01$, 2.8\% degradation) impacts performance more than insufficient decoder supervision, confirming that quantity-aware feature learning at the encoder stage is critical for effective density prediction.

\subsection{Visualization}
\begin{figure}[ht]
	\centering
	\includegraphics[width=0.47\textwidth]{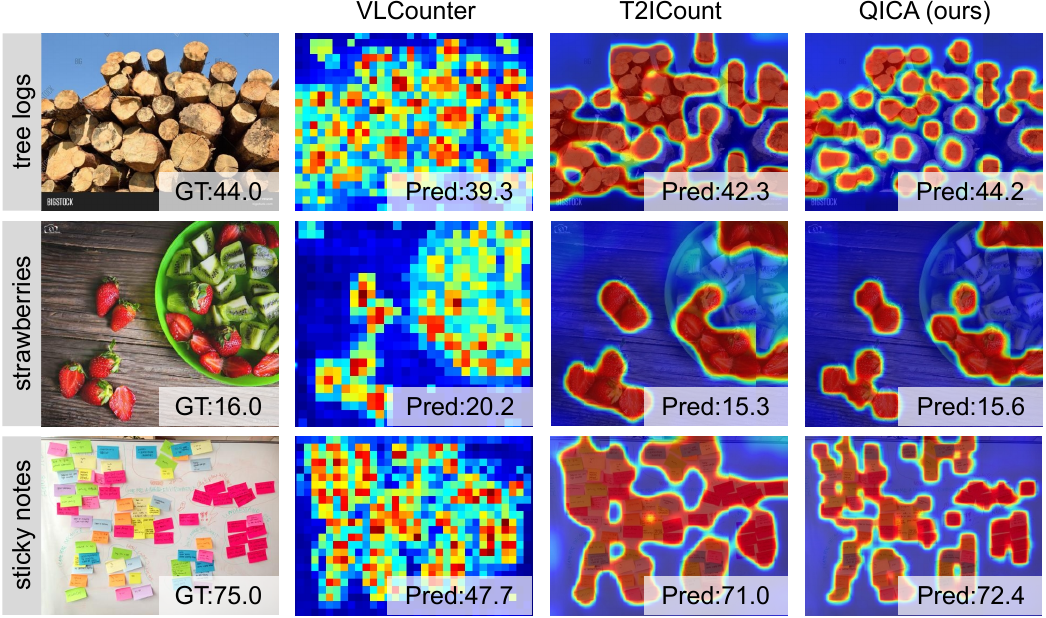} % LaTeX 自带示例图
	\caption{
		Qualitative comparison of QICA with VLCounter \cite{kang2024vlcounter} and T2ICount \cite{qian2025t2icount}. Our text-image `similarity map' exhibits reduced noise and more precise object delineation, which results a more accurate density estimation.
	}
	\label{fig6}
	\vspace{-0.2cm}
\end{figure}

Figure \ref{fig6} compares QICA with VLCounter \cite{kang2024vlcounter} and T2ICount \cite{qian2025t2icount} across diverse categories. 
On tree logs with irregular shapes, QICA  produces smooth and spatially coherent density predictions, while VLCounter shows fragmented responses and T2ICount exhibits less precise localization. 
For strawberries featuring similar appearance and clustered arrangement, QICA and T2ICount achieve accurate counts near GT, whereas VLCounter overestimates due to response leakage. 
On sticky notes with high density and color variation, QICA approaches GT, outperforming VLCounter which severely undercounts. 
The visualization demonstrates that QICA's cost aggregation decoder produces more refined spatial distributions with better boundary delineation, validating its effectiveness in preserving quantity-aware localization across challenging scenarios.

\begin{figure}[ht] % [ht] 是浮动选项，尝试将图片放在“这里(here)”或“顶部(top)”
	\centering % 整个 figure 环境居中
	
	% --- 第一个子图 ---
	\begin{subfigure}[b]{0.236\textwidth}
		\centering
		\includegraphics[width=\textwidth]{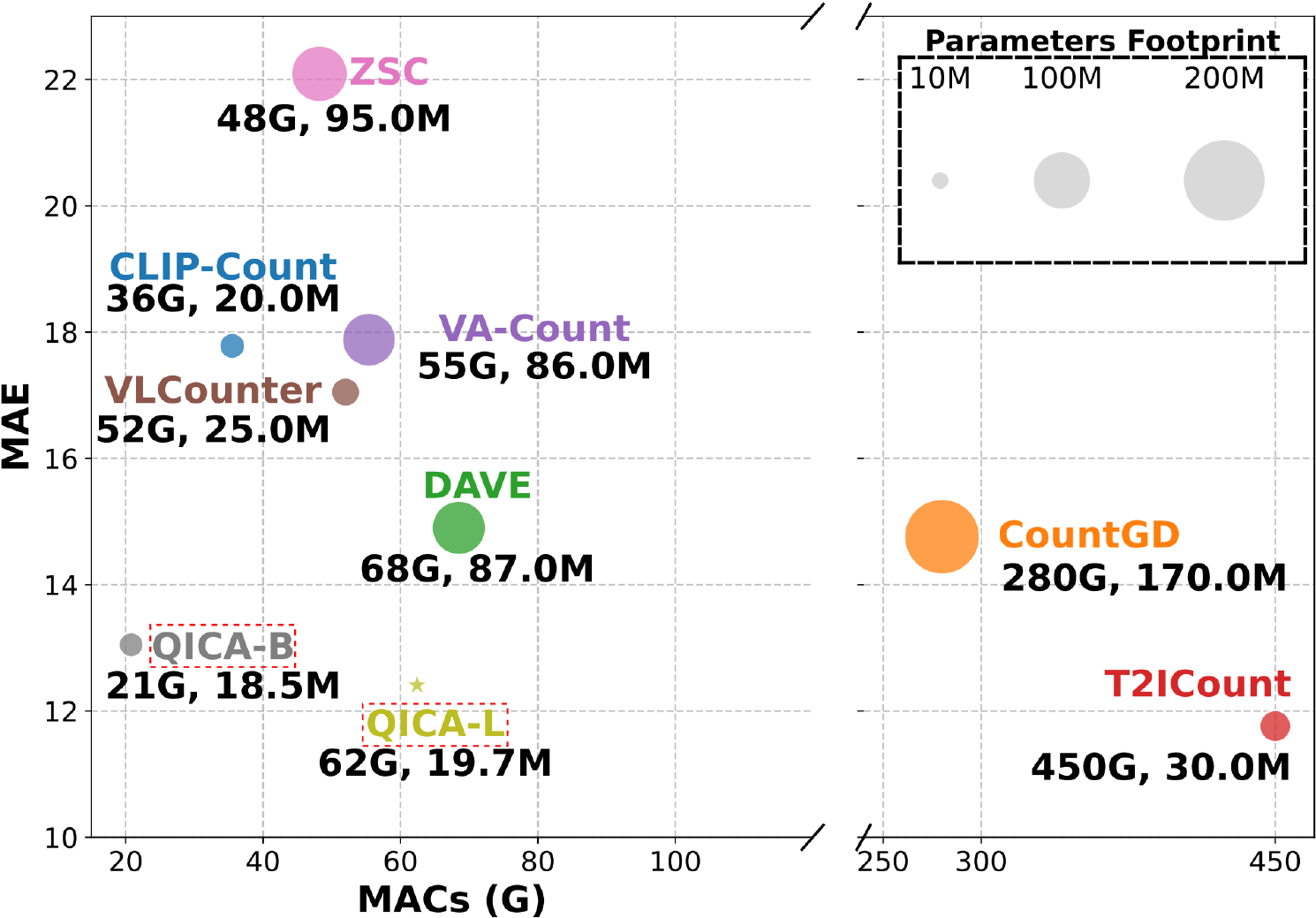} % <-- 替换为您的图片路径
		\caption{}
		\label{xiaolv1}
	\end{subfigure}
	\hfill % 这是一个“弹簧”，在两个子图之间添加水平间距
	% --- 第二个子图 ---
	\begin{subfigure}[b]{0.236\textwidth}
		\centering
		\includegraphics[width=\textwidth]{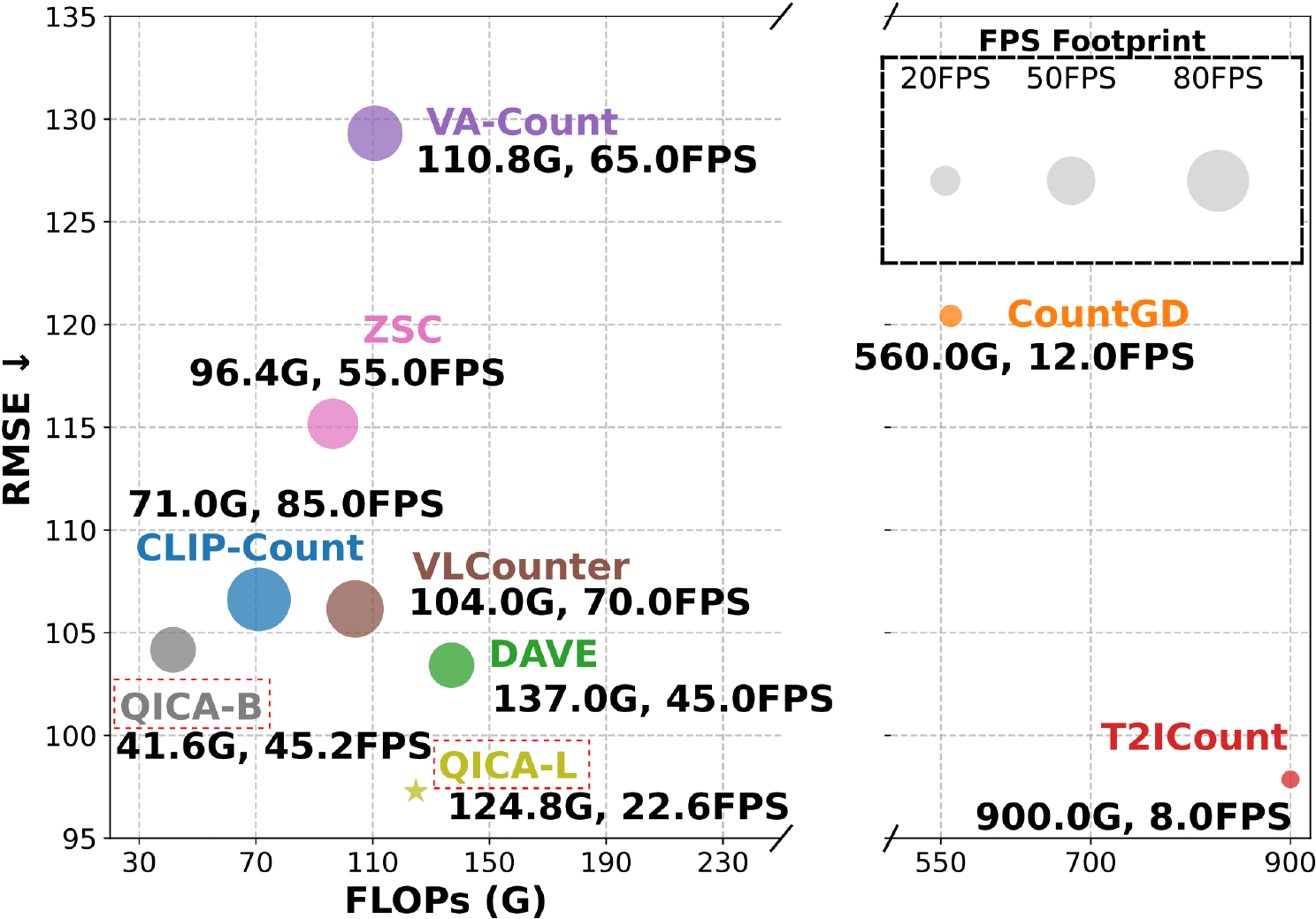} % <-- 替换为您的图片路径
		\caption{}
		\label{xiaolv2}
	\end{subfigure}
	 % 可选：在子图和主标题之间添加一点垂直空间
	
	% --- 主标题 ---
	\caption{
Efficiency analysis of QICA compared to SOTA methods. 
(a) Relationship between trainable parameters, MAE performance, and computational cost represented. 
(b) Comprehensive efficiency metrics including FLOPs and inference speed (FPS). \textcolor{Gold}{\ding{72}} represents QICA-L (ViT-L/14) while \scalebox{1.5}{{\color{gray}\textbullet}} is QICA-B (ViT-B/16).
		}
	\label{fig7}
		\vspace{-0.2cm}
\end{figure}

\subsection{Efficiency Analysis}

Figure \ref{fig7} presents comprehensive efficiency comparison on NVIDIA A800 GPU. QICA demonstrates exceptional parameter efficiency with only 18.5M/19.7M trainable parameters versus 25M-170M for existing methods, achieved through frozen encoders with learnable prompt adaptation. 
QICA-B/16 delivers 20.8G MACs and 45.2 FPS while achieving superior accuracy compared to VLCounter. 
The ViT-L/14 variant balances computational cost with enhanced performance and competitive speed. 
Both configurations significantly outperform heavyweight approaches like CountGD and T2ICount, positioning QICA in the optimal efficiency-performance region for practical deployment.
More detailed information are in Appendix.

\section{Conclusion}

We present QICA, a novel framework for ZSOC that addresses critical limitations regarding fine-grained quantity awareness and spatial insensitivity in existing approaches. 
Our SPS enables effective joint adaptation through explicit numerical conditioning, instilling the model with precise quantity perception via bidirectional gradient coupling. 
Simultaneously, the CAD overcomes spatial insensitivity by refining similarity maps. 
This design effectively prevents feature space distortion during adaptation while recovering the fine-grained spatial structure necessary for precise counting. 
Additionally, the ($\mathcal{L}_{MQA}$) enforces numerical consistency across encoder and decoder stages through ranking constraints. 
Extensive experiments demonstrate competitive performance on FSC-147, with cross-dataset evaluation on CARPK and ShanghaiTech-A validating strong zero-shot generalization to unseen domains.

\noindent
\textbf{Limitation and Future Work:}
QICA relies on discrete quantity hypotheses and focuses on single-category counting, which limits its applicability to continuous quantity relationships and multi-category scenarios. 
We will explore continuous quantity representations and simultaneous multi-category counting to broaden practical utility in the future.

\section*{Acknowledgements}

This work was supported in part by grants from the National Natural Science Foundation of China (62576284 \& 62306241), and in part by grants from the Innovation Foundation for Doctor Dissertation of Northwestern Polytechnical University (No.CX2025109).

\newpage
{
    \small
    \bibliographystyle{ieeenat_fullname}
    \bibliography{main}
}


\end{document}